%% file: main.tex
\documentclass[10pt,twocolumn,letterpaper]{article}

\usepackage[pagenumbers]{cvpr} 

\usepackage{color}
\usepackage{amsfonts}
\usepackage{amsmath}
\usepackage{amssymb}
\usepackage{colortbl}
\usepackage{eqparbox}
\usepackage{arydshln}
\usepackage{multirow}
\usepackage{algorithm,algpseudocode}
\usepackage{stackengine}
\usepackage{soul}
\usepackage[dvipsnames]{xcolor}
\newcommand\xrowht[2][0]{\addstackgap[.5\dimexpr#2\relax]{\vphantom{#1}}}

\definecolor{cvprblue}{rgb}{0.21,0.49,0.74}
\usepackage[pagebackref,breaklinks,colorlinks,allcolors=cvprblue]{hyperref}

\def\paperID{****} 
\def\confName{CVPR}
\def\confYear{2026}

\title{Understanding and Mitigating Hallucinations  in \\ Multimodal Chain-of-Thought Models}

\author{Ji Ma$^{1,2}$, Wei Suo$^{1,2}$\footnotemark[3] ,
 Peng Wang$^{1,2}$, Yanning Zhang$^{1,2}$
\\
$^1$School of Computer Science and Ningbo Institute, Northwestern Polytechnical University, China. \\
$^2$ National Engineering Laboratory for Integrated
Aero-Space-Ground-Ocean \\
Big Data Application
Technology, China. \\
{\tt\small maji@mail.nwpu.edu.cn 
\tt\small \{suowei,peng.wang,ynzhang\}@nwpu.edu.cn
}}

\begin{document}
\maketitle

\renewcommand{\thefootnote}{\fnsymbol{footnote}} 
\footnotetext[3]{Corresponding author.}

\input{sec/0_abstract}    
\input{sec/1_intro_backup}

\input{sec/2_related_work}
\input{sec/3_preliminary}

\input{sec/4_method}

\input{sec/5_experiments}

\input{sec/6_conclusion}

\clearpage

{
    \small
    \bibliographystyle{ieeenat_fullname}
    \bibliography{main}
}

\end{document}

%% file: sec/0_abstract.tex
\begin{abstract}

Multimodal Chain-of-Thought (MCoT) models
have demonstrated impressive capability in complex visual reasoning tasks.
Unfortunately, recent studies reveal that they suffer from severe hallucination problems due to diminished visual attention during the generation process.
However, visual attention decay is a well-studied problem in Large Vision-Language Models (LVLMs). 
Considering the fundamental differences in reasoning processes between MCoT models and traditional LVLMs, we raise a basic question: Whether MCoT models have unique causes of hallucinations?
To answer this question, we systematically investigate the hallucination patterns of MCoT models and find that fabricated texts are primarily generated in associative reasoning steps, which we term divergent thinking. 
Leveraging these insights, we introduce a simple yet effective strategy that can effectively localize divergent thinking steps and intervene in the decoding process to mitigate hallucinations. Extensive experiments show that our method outperforms existing methods by a large margin. More importantly, our proposed method can be conveniently integrated with other hallucination mitigation methods and further boost their performance. The code is publicly available at \href{https://github.com/ASGO-MM/MCoT-hallucination}{https://github.com/ASGO-MM/MCoT-hallucination}.

\end{abstract}

%% file: sec/1_intro_backup.tex
\section{Introduction}
\label{sec:intro}

The advent of Chain-of-Thought (CoT) has enabled powerful reasoning models (\emph{e.g.,} OpenAI-o1/o3~\cite{o1} and Deepseek-R1~\cite{Deepseek-r1}), which have significantly extended the capabilities of Large Language Models (LLMs) in complex problem planning and solving~\cite{yang2025qwen3,face2025open_r1,Gpt-4,Deepseek-v3}. Recently, the multimodal community has transferred this capacity to Large Vision-Language Models (LVLMs)~\cite{qwen-2.5-vl,llava,ma2025short,suo2025pruning} by reasoning about images in a pure-text form~\cite{Mammoth-vl,peng2025skywork,tanreason,R1-onevision} or thinking with image elements (\textit{e.g.,} bounding boxes~\cite{grit} and cropped images~\cite{pixelreasoner}). Different from traditional methods that can only provide final answers, the Multimodal Chain-of-Thought (MCoT) paradigm enables models to interpret images via textual reasoning rationales~\cite{mcot_survey} and facilitate the cross-modal interaction by grounding reasoning chains in images~\cite{think_survey,su2025openthinkimg}.

\begin{figure}[t]
  \centering
  \includegraphics[width=1\linewidth]{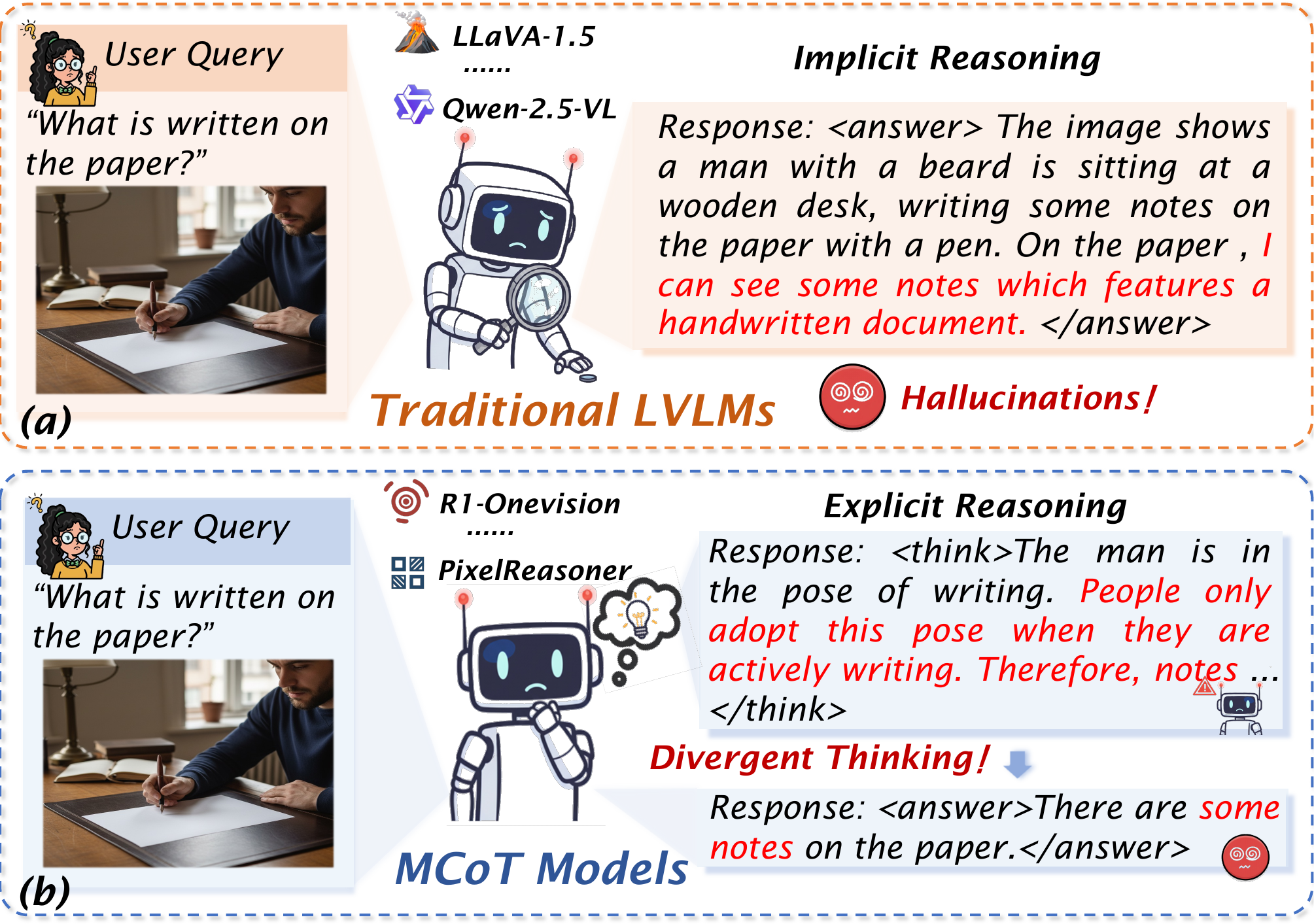}
  \caption{
  Visualization of different reasoning paradigms. (a) Traditional LVLMs apply an implicit reasoning paradigm. (b) MCoT models reason explicitly and answers are influenced by the thinking process.
  }
  \label{fig:motivation}
\end{figure}

Despite these advancements, most existing MCoT efforts~\cite{Mammoth-vl,peng2025skywork,R1-onevision,grit,pixelreasoner,Visual-rft} have focused on improving \textit{reasoning} capability, often overlooking the \textit{perception} aspect. Until recently, several studies~\cite{more_think_less_see,more_thought_less_acc,look_back} revealed that
MCoT models suffer from severe hallucinations, \emph{i.e.,} incorrect visual perceptions that cause textual answers to conflict with the visual evidence. 
These works attribute hallucinations to the diminished visual
attention during MCoT model's generation process~\cite{more_think_less_see,look_back,more_thought_less_acc}.
While plausible, visual
attention decay has been a long-standing problem for hallucination in LVLMs~\cite{liu2024paying,zheng2025lvlms,huang2024opera}, 
and numerous studies have investigated this issue~\cite{liu2024paying,zheng2025lvlms,huang2024opera,tang2025seeing,prabhakaran2025vade,jiang2025devils}.
As shown in Fig.~\ref{fig:motivation}, considering the fundamental differences in reasoning processes between traditional LVLMs (\emph{i.e.,} implicit reasoning) and MCoT models (\emph{i.e.,} explicit reasoning), we raise a basic question: \textbf{\textit{Whether MCoT models have unique causes of hallucinations?}}

To answer the question, we examine the hallucination patterns of MCoT models across both thinking and answering stages.
Specifically, by examining co-occurrence relationships and attention distributions, we find that hallucinations stem from the model's thinking and are propagated to answers through the self-attention mechanism.  
Moreover, further fine-grained analyses on reasoning chains reveal that fabricated texts are primarily generated in associative reasoning steps, which we term \textit{divergent thinking}.
As shown in Fig.~\ref{fig:motivation}~(b), MCoT models would 
rely on their internal reasoning ability rather than actual visual evidence during this process, thereby leading to hallucinations.

The above findings suggest that \textit{divergent thinking} is a key factor for hallucinations in MCoT models.
Thus, a natural idea is to identify this thinking mode and modulate the model's reasoning to mitigate hallucinations.
However, without well-defined labels, identifying the divergent thinking steps for MCoT models is difficult. Moreover, 
how to rectify the divergent thinking process also remains an unsolved challenge.

To tackle these problems, we
introduce an entropy-based score \textit{visual entropy} to measure model's internal confidence towards images.
The underlying hypothesis is that MCoT models prioritize reasoning over perception during divergent thinking steps, which may lead to higher uncertainty in images.
Through comprehensive experiments, we validate this hypothesis and show the effectiveness of this score in localizing MCoT model's divergent thinking steps (more discussion in §\ref{sec:application_1}).
Building on \textit{visual entropy}, we further design a simple yet effective strategy to mitigate hallucinations by adaptively adjusting the decoding process. 
Benefiting from the above designs, our method not only effectively reduces hallucinated contents, but also improves perceptual and reasoning capabilities. 
More importantly, our strategy offers mechanistic understandings of hallucination patterns and can integrate with various traditional hallucination mitigation methods (\textit{e.g.,} DoLa~\cite{chuang2023dola} and VCD~\cite{vcd}) to boost their utility for MCoT models.
In summary, our contributions are summarized as follows:

1) We systematically analyze hallucination causes in MCoT models and uncover that \textit{divergent thinking} is a key factor for hallucination emergence.

2) We connect model's external hallucination behaviors to internal \textit{visual entropy} and demonstrate that this strategy can be used to effectively locate and mitigate hallucinations.

3) Extensive experiments show that our method not only mitigates hallucinations in MCoT models but also boosts their general perception and reasoning capabilities.

%% file: sec/2_related_work.tex
\section{Related Work}
\label{sec:related_work}

\subsection{Multimodal Chain-of-Thought models}

The success in text-only reasoning inspired the development of Multimodal Chain-of-Thought (MCoT) for Large Vision-Language Models (LVLMs)~\cite{llava,suo2026semi,ma2024c3}, resulting in MCoT models~\cite{Zhang2023mcot, mcot_survey,Visual-rft,think_survey,R1-onevision,grit,pixelreasoner}. MCoT extends the CoT principle by teaching LVLMs to generate reasoning chains based on images in pure-text form~\cite{R1-onevision,mcot_survey,Zhang2023mcot,zhao2025cot} or image-based elements~\cite{think_survey,su2025openthinkimg,pixelreasoner,grit}, allowing models to tackle complex multimodal tasks such as complex spatial reasoning~\cite{liimagine,wang2025visuothink,chen2024spatialvlm} and embodied task planning~\cite{liang2025memory,zhao2025cot,wangrobogen}. 
While MCoT models excel at multimodal reasoning, they are prone to the hallucination issue, \textit{i.e.,} generating textual answers that deviate from the visual input~\cite{more_think_less_see,more_thought_less_acc}. In this paper, we focus on alleviating the hallucination issue within the MCoT framework.

\subsection{Hallucinations in MCoT models}
Hallucination has been a long-standing problem in LVLMs~\cite{chair,pope,mmhal}. 
The causes are multi-faceted, often attributed to factors like visual attention decay~\cite{liu2024paying,jiang2025devils,huang2024opera} and language prior~\cite{he2025evaluating,vcd,pope}.
While various efforts~\cite{vcd,chuang2023dola,yang2025nullu,memvr,jiang2025devils,liu2024paying} have been made to mitigate hallucinations in LVLMs, related studies are still limited for MCoT models.
Recently, Liu et al.~\cite{more_think_less_see} find that as the reasoning chain lengthens, the MCoT model's attention to the visual input progressively degrades, causing increased hallucinations. Tian et al.~\cite{more_thought_less_acc} attribute this problem to the “dual nature of reasoning”, identifying a trade-off where longer reasoning causes a marked decrease in visual attention, 
leading to more hallucinated contents.
 Different from these approaches, our work systematically investigates hallucination patterns in MCoT models and uncovers that \textit{divergent thinking} is a key cause of this issue. 
 

%% file: sec/3_preliminary.tex
\section{Understanding Hallucinations in MCoT}
\label{sec:3}

This paper systematically examines hallucinations in Multimodal Chain-of-Thought (MCoT) models, providing in-depth insights into the causes of this issue.
We start by outlining the experimental setup of our case studies. 
Then, we conduct experiments to explore the key question: What is the unique cause of hallucination in MCoT models?

\begin{figure}[t]
  \centering
  \includegraphics[width=1\linewidth]{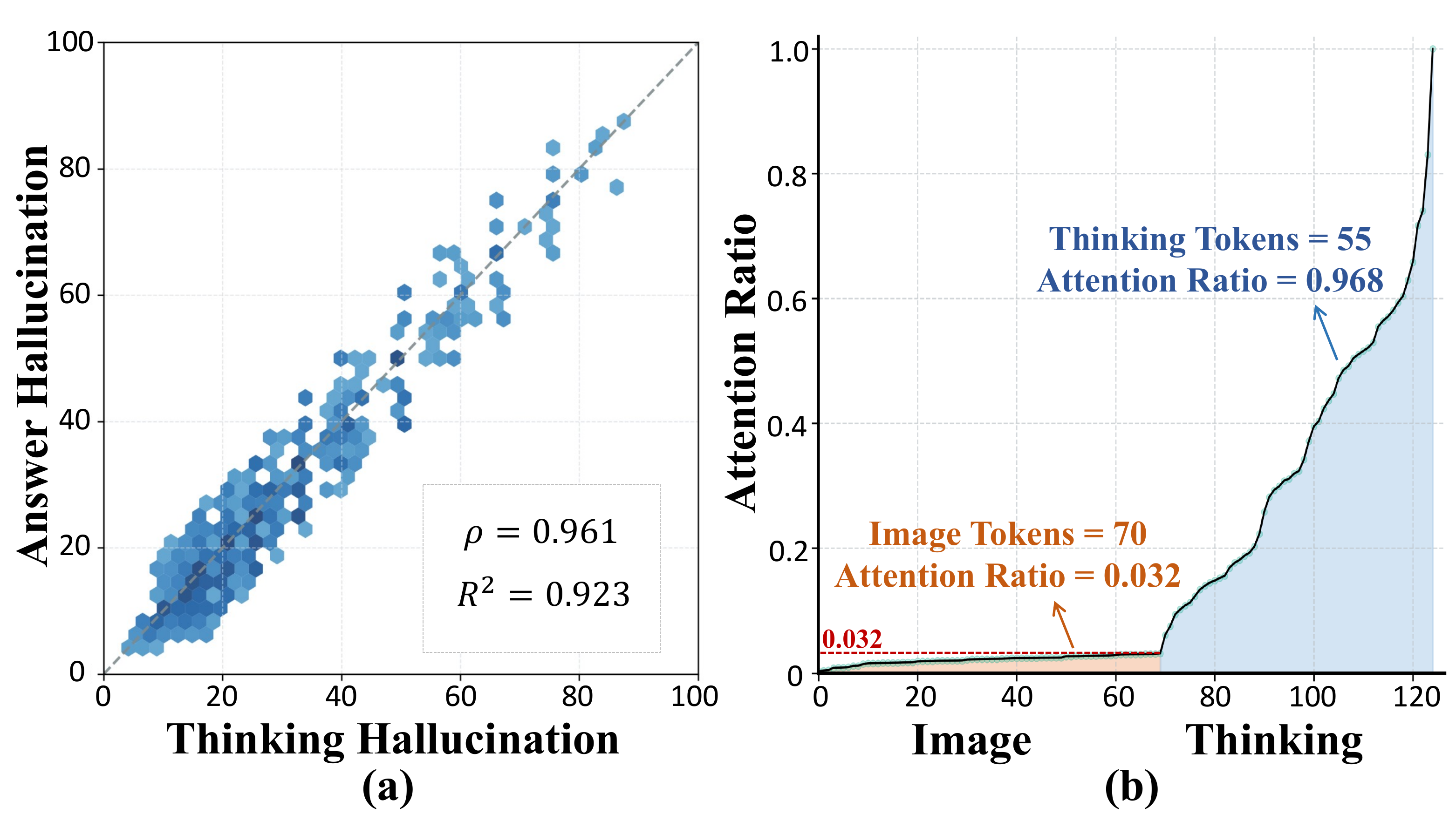}
  \caption{(a) Co-occurrence relationship of hallucinations in the MCoT model's thinking and answering. We find that hallucinations in thinking and answering are positively correlated. (b) Further attention visualization reveals that this phenomenon is caused by the MCoT model's attention bias towards its thinking process when generating final answers.}
  \label{fig:pre_1}
\end{figure}

\subsection{Preliminary}
\label{sec:preliminary}

Given the visual input $v$ and a user's query $q$, MCoT models perform multi-hop reasoning to better address complex multimodal problems. The model outputs can be divided into reasoning and answers by the special tokens $<$\textit{think}$>$ and $<$\textit{answer}$>$. 
Specifically, at each generation step $t$, the auto-regressive decoding can be formalized as follows:
\begin{equation}
    y_t\sim p_{t}(\cdot|v,q,y_{<t}),
\label{eq:1}
\end{equation}
where $p_t$ is the probability distribution over the vocabulary $\mathcal{V}$.
Leveraging reasoning chains, MCoT models demonstrate impressive multimodal reasoning capabilities across a wide range of tasks. However, these models suffer from severe hallucination problems, \textit{i.e.,} fabricated textual answers that conflict with visual content~\cite{more_think_less_see,more_thought_less_acc}.
To understand the hallucination issue in MCoT models, we conduct experiments to explore: 1) whether hallucinations in MCoT models' answers originate from their thinking process; 2) if so, what hallucination patterns occur in the reasoning chains.

\subsection{Experimental Setup for Case Study}
\label{sec:Experimental Setup for Case Study}

In this section, we employ the widely used Object HalBench~\cite{chair} and POPE~\cite{pope} benchmarks to conduct case studies. 
For Object HalBench, we follow previous work~\cite{jiang2025devils} and use the prompt “\textit{Please help me describe the image in detail.}”. The hallucination ratio is measured with the default CHAIR score~\cite{chair}.
For POPE, following previous works~\cite{yang2025nullu}, we use the prompt “\textit{Please answer this question in one word.}” and evaluate with the accuracy.
Meanwhile, regarding model selection, we mainly focus on the R1-Onevision-7B~\cite{R1-onevision} in this section. 
To ensure that our findings are generalizable, all experiments are repeated three times using different random seeds. 

\subsection{Finding 1: Thinking Propagates Hallucinations to Answers}
\label{sec:finding1}

In this section, we conduct two types of experiments (\emph{i.e.,} examining co-occurrence relationships and attention distributions) to explore whether hallucinations in the final answers stem from the reasoning process.
For the co-occurrence analysis, we first use the $<$\textit{think}$>$ segments to separate the model's thinking from answers. Then, standard metrics from Sec.~\ref{sec:Experimental Setup for Case Study} are applied to calculate a paired hallucination ratio (thinking \emph{vs.} answering) for each sample. As shown in Fig.~\ref{fig:pre_1}~(a), this analysis reveals a strong positive correlation between hallucination ratios of the thinking and answering stages (with $\rho$ $>$ $0.96$ and $R^2>0.92$). Here, $\rho$ reflects the linear dependency and $R^2$ measures the correlation strength.

The above phenomenon indicates that fabricated answers are strongly correlated with hallucinations in the thinking process. We hypothesize that the underlying reason lies in the model's self-attention mechanism~\cite{vaswani2017attention}, which may propagate unfaithful information from preceding reasoning steps to generate hallucinated answers.
To validate this, we visualize the attention distribution of answer tokens on the image and thinking tokens. 
Note that the system tokens and instruction tokens are omitted for simplicity.
As shown in Fig.~\ref{fig:pre_1}~(b), we use “orange” and “blue” colors to denote the attention ratio received by the image and thinking parts. From this figure, we observe that the model's attention is heavily biased towards the thinking process, with less than $0.04$ allocated to the image.
Based on the above results, we conclude that \textbf{\textit{hallucinations in MCoT models originate from the thinking process and these fabricated contents are further propagated to answers through the attention mechanism.}}

\subsection{Finding 2: Divergent Thinking Is a Unique Cause of Hallucinations}
\label{sec:finding2}

The above analyses reveal that hallucinations in MCoT models stem from their thinking process. In this section, we delve deeper into the model's reasoning chains to provide a fine-grained investigation of the hallucination patterns.

\begin{figure}[t]
  \centering
  \includegraphics[width=1\linewidth]{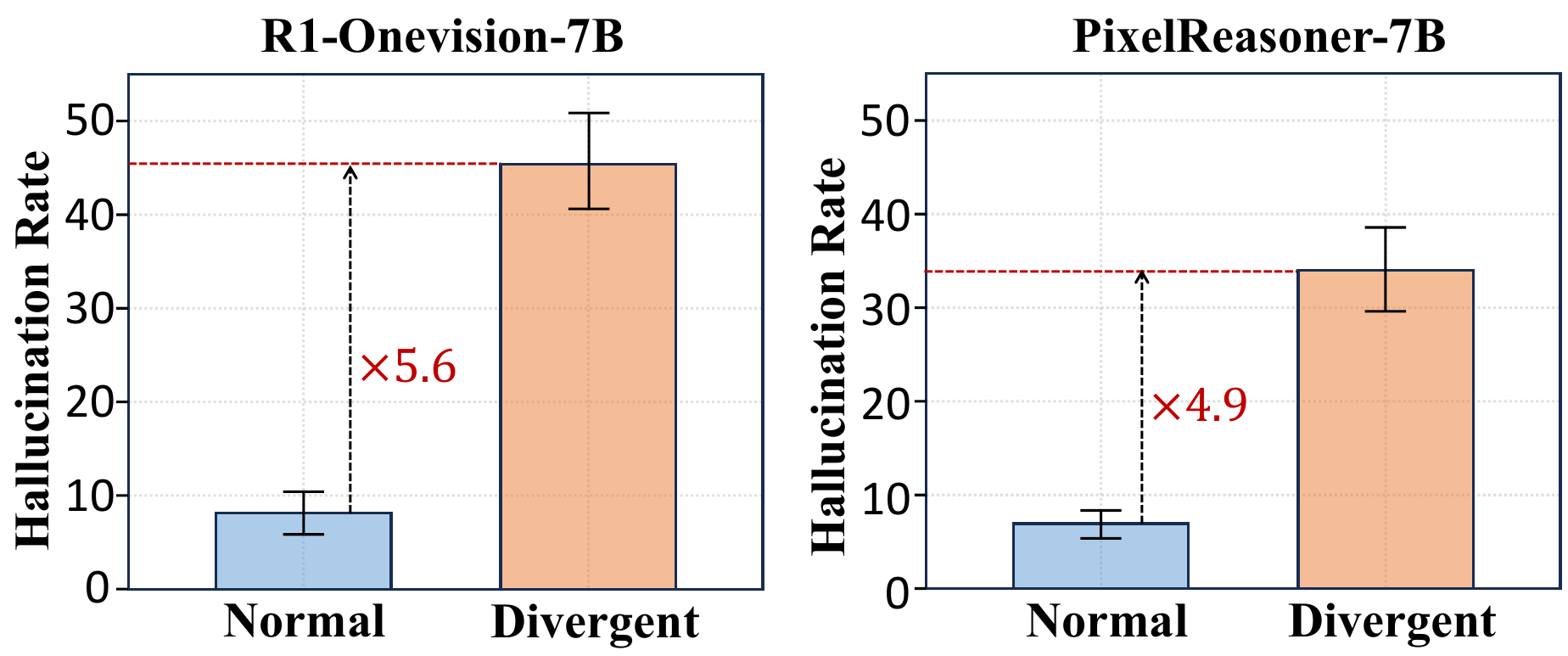}
  \caption{Hallucination patterns in the MCoT model's thinking process. We find that models are prone to hallucinations during associative reasoning steps, termed as \textit{divergent thinking}. In this thinking mode, MCoT models exhibit $\sim$5 times more hallucinations compared to normal thinking.}
  \label{fig:thinking_mode}
\end{figure}

Research in cognitive science has shown that humans would jointly apply \textit{divergent thinking} and \textit{normal thinking} to handle complex reasoning tasks~\cite{congnitive_science}.
Inspired by this, we attempt to divide the MCoT reasoning process into several segments (each labeled as \textit{divergent} or \textit{normal}) based on their associative strengths.
For instance, as shown in Fig.~\ref{fig:motivation}~(b), segments highlighted in red are considered divergent thinking. The classification is performed semi-automatically: we first use GPT-5~\cite{openai_gpt5_systemcard} to obtain initial annotations, and three graduate-level annotators then verify them. 
After decomposing the reasoning chains into labeled thinking segments (\textit{normal} or \textit{divergent}), we calculate the hallucination rate for each segment and then average these scores by their category. As shown in Fig.~\ref{fig:thinking_mode}, the results reveal that divergent thinking steps are more prone to hallucinations, exhibiting the hallucination rate $\sim 5$ times higher than normal thinking. Therefore, we conclude that \textbf{\textit{fabricated texts are not uniformly generated in model's reasoning and divergent thinking is a unique cause of hallucinations in MCoT models. }}

%% file: sec/4_method.tex
\section{Mitigating Hallucinations in MCoT Models}
\label{sec:4}

From the analyses in Sec.~\ref{sec:3}, it can be found that \textit{divergent thinking} is a key factor of hallucinations in MCoT models. Therefore, a natural idea is to recognize these thinking steps and adjust them to improve faithfulness. 
However, identifying the divergent thinking mode is difficult due to the lack of well-defined labels. Moreover, how to rectify the specific thinking process presents another important challenge.

To tackle these problems, we propose an entropy-based score \textit{visual entropy} to measure the MCoT model's internal confidence in images. 
The underlying hypothesis is that MCoT models exhibit an inner divergent thinking pattern that prioritizes reasoning over perception, which may result in higher uncertainty about visual input.
In this section, we empirically validate whether \textit{visual entropy} can serve as a reliable indicator for the model's divergent thinking steps, and how we can utilize this score to mitigate hallucinations.
Next,  we would introduce the formal definition of the proposed \textit{visual entropy} in detail.

\subsection{Visual Entropy}

Given the visual inputs $v= [v_1,...,v_m]$, we first obtain the hidden states $\mathbf{v}^L\in \mathbb{R}^{d\times m}$ by feeding them into the MCoT model. Here, $L$ and $d$ are the model's layer number and hidden dimension, respectively. Then, we apply the language head and \textit{softmax} to map these representations to probability distributions over the vocabulary $\mathcal{V}$, obtaining $\mathbf{p}_{v} \in \mathbb{R}^{|\mathcal{V}|\times m}$.
Finally, for the predictive token $y_t$, we extract the visual activation probabilities $\mathbf{p}_{v}(y_t)\in \mathbb{R}^{m}$ using its index in the vocabulary $\mathcal{V}$. 
Here, the $\mathbf{p}_{v}(y_t)$ denotes
how likely $y_t$ can be generated by visual tokens~\cite{visual_inte}.
Based on $\mathbf{p}_{v}(y_t)$, the \textit{visual entropy} is formulated as:
\begin{equation}
E(y_t,v) = - \frac{\sum_{i=1}^{m} p_{v,i}(y_t) \log(p_{v,i}(y_t))}{\log m},
\label{eq:2}
\end{equation}
where $p_{v,i}(y_t)$ is the $i$-th element of $\mathbf{p}_{v}(y_t)$, $m$ is the sequence length of the visual inputs, and $\log m$ is used to normalize the entropy to $[0, 1]$.

\begin{figure}[t]
  \centering
  \includegraphics[width=1\linewidth]{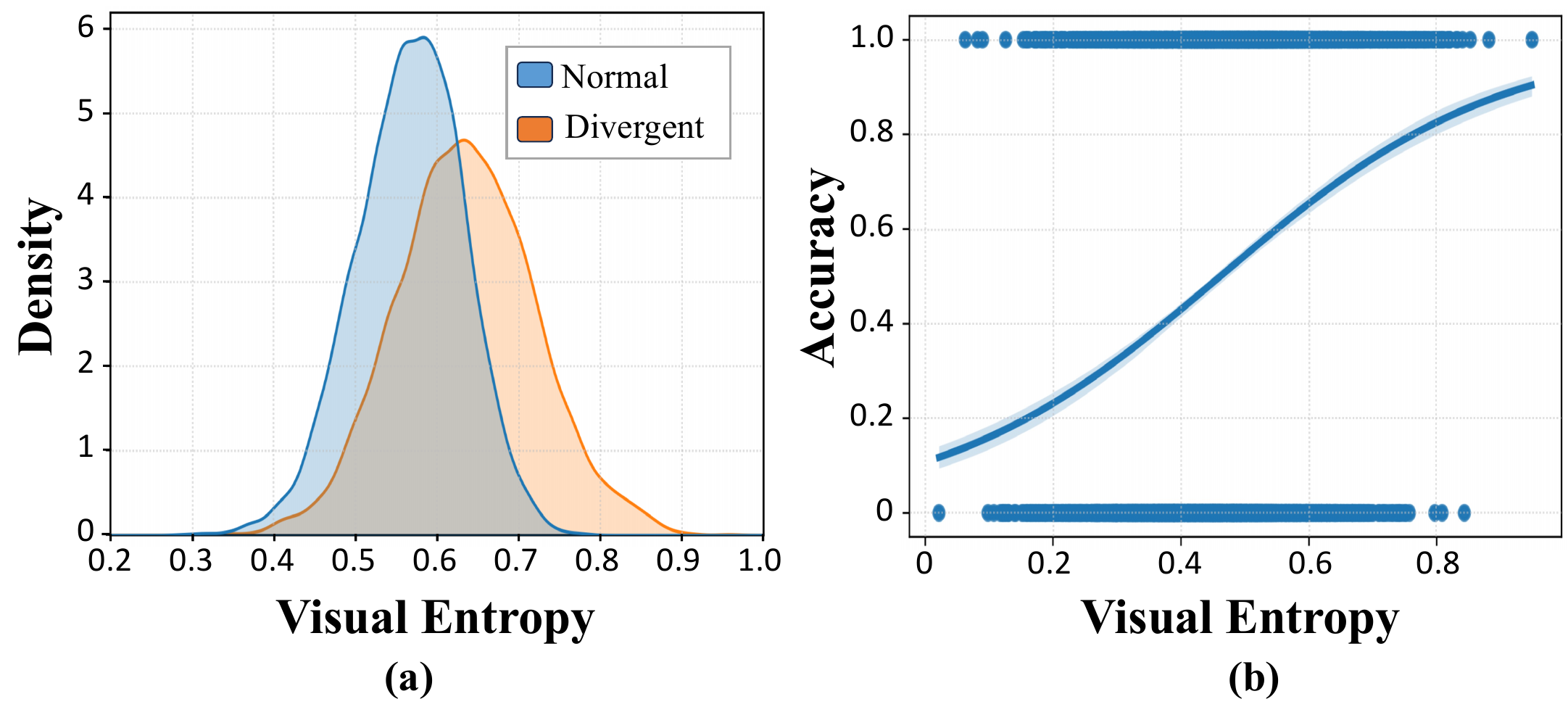}
  \caption{
  (a) \textit{Visual entropy} across different thinking modes. The results show that models exhibit higher values when engaging in divergent thinking.
(b) 
 Predicting the divergent thinking mode using \textit{visual entropy}. The logistic curve demonstrates that high visual entropy reliably predicts the divergent thinking steps.
  }
  \label{fig:localize}
\end{figure}

\subsection{Application 1: Visual Entropy Can Identify Divergent Thinking}
\label{sec:application_1}
After obtaining \textit{visual entropy}, we investigate whether this score can distinguish between different thinking modes. 
Using the labels from Sec.~\ref{sec:finding2} (which classify reasoning segments as \textit{normal} or \textit{divergent}), we compute the entropy value for each segment and then average these scores by category.
As shown in Fig.~\ref{fig:localize}~(a), the entropy values for normal thinking are evidently lower than divergent thinking, suggesting that \textit{visual entropy} can be utilized as an effective indicator.
We further validate this by applying a simple logistic regression model~\cite{lavalley2008logistic} to classify thinking modes. As illustrated in Fig.~\ref{fig:localize}~(b), 
the logistic curve shows that high visual entropy reliably predicts the divergent thinking segments.
Additionally, we follow previous work~\cite{gema2024decore} and  compute McFadden’s pseudo-$R^2$~\cite{mcfadden1972conditional}, which exceeds $0.9$, demonstrating that divergent thinking is strongly correlated with entropy values. Based on these results, it can be found that \textit{visual entropy} is highly predictive of divergent thinking steps, thereby offering an effective approach to identify MCoT model's reasoning modes. 
x

\begin{figure}[t]
  \centering
  \includegraphics[width=1\linewidth]{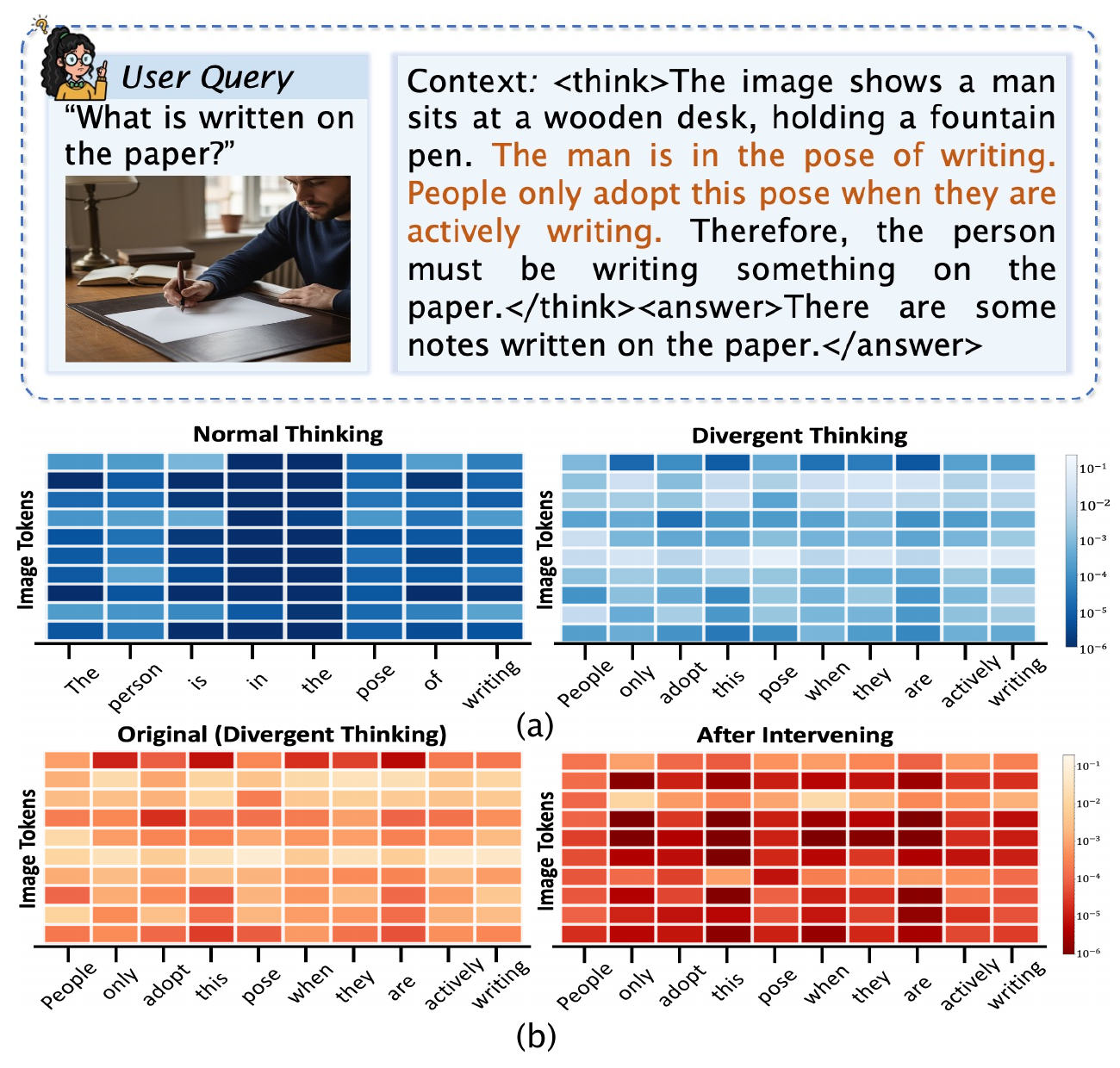}
  \caption{Visualization of \textit{visual entropy}. We randomly choose ten image tokens for clarity, and dark colors denote \textit{low} entropy values. (a) Compared to normal thinking, entropy values are significantly higher in divergent thinking steps. (b) After applying our method, \textit{visual entropy} shows an evident decline.}
  \label{fig:pre_qual}
\end{figure}

\subsection{Application 2: Visual Entropy Can Mitigate Hallucinations}
\label{sec:application_2}
The above analyses demonstrate that \textit{visual entropy} correlates with MCoT models' divergent thinking, and this mode exhibits higher entropy values compared to normal reasoning. Based on this, we further examine whether we can adjust the divergent thinking and mitigate hallucinations via \textit{visual entropy}. 
Inspired by constrained decoding approaches~\cite{constrained_1,constrained_2,in-context-sharp}, we incorporate \textit{visual entropy} as additional constraints to MCoT model's decoding process. For the $t$-th divergent thinking step, we revise the model's decoding as follows:
\begin{equation}
    \hat{p}_{t}(\cdot|v,q,y_{<t})= p_{t}(\cdot|v,q,y_{<t})\cdot e^{-\alpha\cdot E(\cdot,v)},
\label{eq:3}
\end{equation}
where $v$, $q$ and $\alpha\in [0,1]$ 
 are visual inputs, user query and intervention degree, respectively. $E(\cdot,v)\in \mathbb{R}^{|\mathcal{V}|}$ is the \textit{visual entropy} for the vocabulary $\mathcal{V}$.
Guided by Eq.~\ref{eq:3}, MCoT models are encouraged to select tokens with lower \textit{visual entropy} and penalize those leading to higher entropy. 

To give an intuitive understanding of this process, we compare the entropy values across three settings: 1) normal thinking; 2) divergent thinking; 3) divergent thinking after applying Eq.~\ref{eq:3}.
As shown in Fig.~\ref{fig:pre_qual}~(a), R1-Onevision-7B~\cite{R1-onevision} exhibits high \textit{visual entropy} during divergent thinking. Moreover, in Fig.~\ref{fig:pre_qual}~(b), entropy values in divergent reasoning steps are decreased after we intervene in the decoding process.
In addition to qualitative analyses, 
we also conduct quantitative experiments in Sec.~\ref{Different Model Settings} and Table.~\ref{tab:different_model_settings}.
From the table, it can be observed that \textit{visual entropy} effectively reduces the hallucination ratios of the backbone model by 30\% (CHAIR$_S$).
These results validate that modulating the decoding process with \textit{visual entropy} can effectively mitigate hallucinations in MCoT models.

\begin{figure}[t]
  \centering
  \includegraphics[width=1\linewidth]{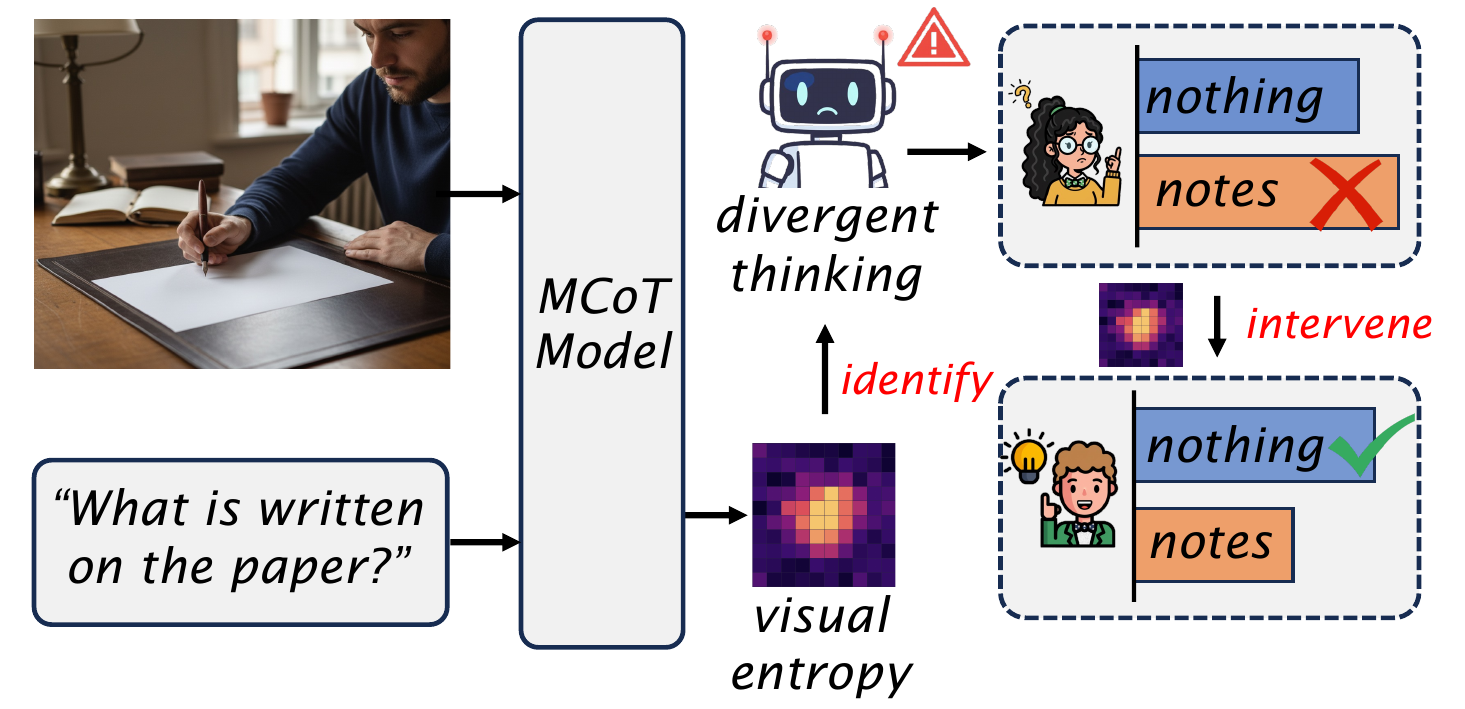}
  \caption{Overview of our method: Given the image and user's query, we first leverage MCoT model's internal \textit{visual entropy} to detect divergent thinking steps. When the model is identified as thinking divergently, we further 
  use this score to dynamically intervene in the decoding process and mitigate hallucinations.}
  \label{fig:model}
\end{figure}

\begin{algorithm}[htbp]
\label{algorithm:1}
\caption{Detailed Steps of Our Strategy}
\label{algorithm:1}
    \begin{algorithmic}[1]
    \State \textbf{Input:} MCoT model $\mathcal{M}$, visual input $v$, user query $q$.
    \State \textbf{Output:} Model Response $y_t$.
    \State Use $\mathcal{M}$ and $v$ to compute $\mathbf{p}_v\in \mathbb{R}^{\mathcal{V}\times m}$ 
    \State For $t$-th reasoning step:
    \State Calculate $E(y_t,v)$ based on $\mathbf{p}_v$ 
    \algorithmiccomment{\textit{visual entropy} of $y_t$}
    \If{$E(y_t,v)>\gamma$} \algorithmiccomment{divergent thinking step}
    \State $\hat{p}_{t}(\cdot|v,q,y_{<t})= p_{t}(\cdot|v,q,y_{<t})\cdot e^{-\alpha\cdot E(\cdot,v)}$ 
    \State Obtain $y_t$ based on $\hat{p}_t$
    \Else \algorithmiccomment{normal thinking step}
    \State Obtain $y_t$ based on $p_t$
    \EndIf
    \end{algorithmic}
\end{algorithm}

\subsection{Overview of Our Strategy}

Building on the applications of \textit{visual entropy}, we propose a training-free strategy to mitigate hallucinations in MCoT models.
Notably, our method operates by adaptively intervening in the decoding process, thereby avoiding 1) the double inference overhead like contrastive decoding~\cite{vcd,chuang2023dola}, and 2) complex internal state manipulations such as hidden representation editing~\cite{yang2025nullu,FlexAC,suohallucination} or visual attention steering~\cite{liu2024paying,jiang2025devils}.
We provide an overview in Fig.~\ref{fig:model}, and detailed steps of our strategy can be found in Algorithm~\ref{algorithm:1}. 
It can be found that our strategy involves two hyperparameters: threshold $\gamma$ and intervention degree $\alpha$. Thus, we explore the effects of these parameters and experimentally show that the hyperparameter choices are model-agnostic, avoiding labor-intensive tuning for new model adaptations (Related discussion in the Sec.~\ref{sec:5.5} and Fig.~\ref{fig:hyper}). Regarding inference efficiency, since the visual token probabilities $\mathbf{p}_v\in \mathbb{R}^{|\mathcal{V}|\times m}$ can be pre-computed during the prefilling stage, the additional computational overhead during decoding is negligible. We also provide experimental validations in Sec.~\ref{sec:5.5} and Fig.~\ref{fig:comparison}~(b).

%% file: sec/5_experiments.tex
\begin{table*}[t]
\centering
\setlength{\tabcolsep}{3.5pt}
\caption{Performance comparison with state-of-the-art methods on several hallucination benchmarks. The best-performing method is highlighted in \textbf{bold}. $\uparrow$ indicates that higher is better, while $\downarrow$ indicates that lower is better.}
\label{tab:main_results}

\begin{tabular}{lcccccccccc}
\toprule
\multicolumn{1}{l}{\multirow{2}{*}{\textbf{Methods}}}  &
\multicolumn{2}{c}{\textbf{Object HalBench}} & \multicolumn{2}{c}{\textbf{MMhalBench}} & \multicolumn{2}{c}{\textbf{POPE} \textit{Random}} & \multicolumn{2}{c}{\textbf{POPE} \textit{Adversarial}} & \multicolumn{2}{c}{\textbf{POPE} \textit{Popular}}      \\    
\multicolumn{1}{c}{}  &  \multicolumn{1}{c}{CHAIR$_S$$\downarrow$}  & \multicolumn{1}{c}{CHAIR$_I$$\downarrow$} & \multicolumn{1}{c}{HalRate$\downarrow$} & \multicolumn{1}{c}{Score$\uparrow$} & \multicolumn{1}{c}{Acc$\uparrow$} & \multicolumn{1}{c}{F1$\uparrow$} & \multicolumn{1}{c}{Acc$\uparrow$}& \multicolumn{1}{c}{F1$\uparrow$ }  & 
\multicolumn{1}{c}{Acc$\uparrow$}& \multicolumn{1}{c}{F1$\uparrow$} \xrowht{6pt} \\ 
\midrule

GRIT-3B~\cite{grit} &  23.8 & 10.5 & 0.51 & 1.94 & 78.1 & 72.2 & 77.5 & 71.8 & 78.6 & 73.1 \\
+ DoLa~\cite{chuang2023dola} &  21.6 & 8.0 & 0.49 & 1.98 & 79.0 & 73.6 & 77.6 & 72.2 & 78.3 & 72.7 \\
+ VCD~\cite{vcd} & 20.4 & 7.9 & 0.50 & 1.97 & 78.9 & 73.4 & 78.1 & 73.0 & 79.0 & 73.7 \\
+ MemVR~\cite{memvr} & 19.4 & 8.1 & 0.46 & 2.01 & 79.5 & 74.3 & 78.7 & 73.8 & 78.9 & 73.6 \\
+ FlexAC~\cite{FlexAC} &  19.2 & 7.4 & 0.48 & 2.04 & 79.8 & 74.9 & 79.0 & 74.1 & 79.6 & 74.7 \\
+ Ours &\textbf{16.0} & \textbf{5.5} & \textbf{0.42} & \textbf{2.15} & \textbf{81.0} & \textbf{76.7} & \textbf{79.8} & \textbf{75.3} & \textbf{80.6} & \textbf{76.1} \\
\midrule

PixelReasoner-7B~\cite{pixelreasoner} & 22.0 & 7.8 & 0.37 & 3.70 & 85.5 & 84.1 & 82.3 & 81.9 & 84.3 & 82.9   \\
+ DoLa~\cite{chuang2023dola} &  20.6 & 7.1 & 0.35 & 3.74 & 85.8 & 84.2 & 82.6 & 81.5 & 85.7 & 84.3 \\
+ VCD~\cite{vcd} & 19.8 & 7.2 & 0.33 & 3.87 & 85.6 & 83.9 & 83.4 & 82.4 & 85.4 & 84.0 \\
+ MemVR~\cite{memvr} & 19.4 & 6.3 & 0.29 & 3.95 & 86.2 & 84.5 & 82.8 & 81.7 & 85.3 & 83.9 \\
+ FlexAC~\cite{FlexAC} & 18.8 & 6.5 & 0.31 & 3.83 & 86.0 & 84.8 & 83.2 & 82.0 & 85.1 & 83.8 \\
+ Ours & \textbf{15.4} & \textbf{5.3} & \textbf{0.26} & \textbf{4.12} & \textbf{87.3} & \textbf{85.6} & \textbf{84.3} & \textbf{83.1} & \textbf{86.5} & \textbf{85.3} \\
\midrule

R1-Onevision-7B~\cite{R1-onevision} &  23.2 & 9.4 & 0.59 & 2.18 & 81.2 & 78.1 & 78.5 & 76.3 & 80.4 & 77.3 \\
+ DoLa~\cite{chuang2023dola} & 21.4 & 9.8 & 0.57 & 2.24 & 82.3 & 79.2 & 79.3 & 77.5 & 80.9 & 77.6 \\
+ VCD~\cite{vcd} & 21.0 & 8.0 & 0.55 & 2.29 & 82.1 & 78.6 & 79.1 & 77.0 & 81.0 & 77.8 \\
+ MemVR~\cite{memvr} & 19.2 & 7.5 & 0.58 & 2.28 & 81.7 & 78.3 & 78.6 & 76.8 & 81.3 & 78.1  \\
+ FlexAC~\cite{FlexAC} & 18.4 & 6.8 & 0.53 & 2.36 & 82.5 & 79.0 & 79.8 & 77.1 & 80.8 & 78.0 \\
+ Ours &\textbf{15.8} & \textbf{5.7} & \textbf{0.50} & \textbf{2.51} & \textbf{83.5} & \textbf{80.3} & \textbf{81.6} & \textbf{78.4} & \textbf{82.2} & \textbf{79.1} \\

\bottomrule
\end{tabular}
\end{table*} 

\section{Experiments}
\label{sec:experiments}

\subsection{Experimental Setting}

\textbf{Datasets and Metrics.} We evaluate our method on three hallucination benchmarks: Object HalBench~\cite{chair}, POPE~\cite{pope} and MMhalBench~\cite{mmhal}.
Moreover, we conduct experiments on three general benchmarks across Perception  (MMStar~\cite{mmstar}) and Reasoning (MathVista~\cite{mathvista} and VSR~\cite{vsr}). 
On Object-HalBench~\cite{chair}, we adopt the default CHAIR$_I$ and CHAIR$_S$ metrics to measure hallucination ratios. For POPE~\cite{pope}, we follow previous work~\cite{suo2025octopus} and report accuracy and F1 score, respectively. We also utilize the GPT-aided evaluation benchmark  MMhalBench~\cite{mmhal} to evaluate the proposed method. For general perception and reasoning benchmarks, we apply standard accuracy as the metric.

\noindent
\textbf{Models and Baselines.}
We evaluate our method on three popular Multimodal Chain-of-Thought (MCoT) models with different model architectures and scales, including GRIT-3B~\cite{grit}, PixelReasoner-7B~\cite{pixelreasoner} and R1-Onevision-7B~\cite{R1-onevision}. To demonstrate the effectiveness of our method, we compare with various SOTA hallucination mitigation methods, including DoLa~\cite{chuang2023dola}, VCD~\cite{vcd}, MemVR~\cite{memvr} and FlexAC~\cite{FlexAC}. 
The discussions on hyperparameter selection can be found in Sec.~\ref{sec:5.5}.

\subsection{Quantitative Evaluation}

\paragraph{Results on Hallucination Benchmarks.}

In Table~\ref{tab:main_results}, we evaluate the effectiveness of our method on three prevalent MCoT models (\textit{i.e.,} GRIT-3B~\cite{grit}, PixelReasoner-7B~\cite{pixelreasoner} and R1-Onevision-7B~\cite{R1-onevision}) across three hallucination benchmarks including Object HalBench~\cite{chair}, MMhalBench~\cite{mmhal} and POPE~\cite{pope}. The results demonstrate that our approach significantly outperforms prior baseline methods.
In particular, for GRIT-3B~\cite{grit}, it can be observed that our method reduces sentence-level hallucinations (CHAIR$_S$) by over 3.0 points compared to the previous SOTA method (\textit{i.e.,} FlexAC~\cite{FlexAC}). Meanwhile, the hallucination reduction on CHAIR$_I$ over the backbone shows a 47.6\% relative improvement. 
Moreover, the effectiveness of our approach generalizes well across different model architectures and scales. For PixelReasoner-7B~\cite{pixelreasoner}, we can see that our method outperforms previous methods by a large margin. Finally, for discriminative tasks (\emph{i.e.,} POPE), our method also demonstrates stronger performance across all three splits.

\noindent
\textbf{Results on Perception and Reasoning Benchmarks.}
To explore whether our method also boosts the general capacity of MCoT models, we further conduct experiments on three perception and reasoning benchmarks, including MathVista~\cite{mathvista}, MMStar~\cite{mmstar}, and VSR~\cite{vsr}.
We use three models (\textit{i.e., }GRIT-3B~\cite{grit}, PixelReasoner-7B~\cite{pixelreasoner}, and R1-Onevision-7B~\cite{R1-onevision}) as the basic models.
In Table~\ref{table:general_benchmarks}, it can be found that our method improves the performance of these models on all benchmarks. Specifically, our approach improves the VSR accuracy for GRIT-3B by 3.2 points. 
For PixelReasoner-7B, the results show that the performance can be further boosted by 1.4 points and 1.9 points on MathVista and VSR, respectively. These results demonstrate that our method not only effectively reduces hallucinated contents, but also improves the perceptual and reasoning capabilities of MCoT models.

\begin{table}
    \centering
    \setlength{\tabcolsep}{5pt}
    \caption{Evaluation on perception and reasoning benchmarks.}
    \begin{tabular}{lccc}
    \toprule 
    \textbf{Method} & \textbf{MathVista} & \textbf{MMStar} & \textbf{VSR} \\
    \midrule
    GRIT-3B~\cite{grit} & 59.8 & 50.7 & 72.9 \\
    + Ours & \textbf{60.9} & \textbf{51.5} & \textbf{76.1} \\
    \midrule
    PixelReasoner-7B~\cite{pixelreasoner} & 62.6 & 58.7 & 82.4 \\
    + Ours & \textbf{64.0} & \textbf{59.4} & \textbf{84.3} \\
    \midrule
    R1-Onevision-7B~\cite{R1-onevision} & 64.1 & 52.8 & 73.4 \\ 
    + Ours & \textbf{65.0} & \textbf{54.2} &  \textbf{74.9} \\
    \midrule
    \end{tabular}
    \label{table:general_benchmarks}
\end{table}

\begin{table}[t]
\centering
\setlength{\tabcolsep}{2.5pt}
\caption{The effects of different settings for our method. “All thinking” denotes intervening in the MCoT model's entire thinking process, while “Normal-only” denotes only modulating the normal thinking steps.}
\label{tab:different_model_settings}
\begin{tabular}{lccc}
\toprule
\multicolumn{1}{l}{\multirow{2}{*}{\textbf{Methods}}}   &   \multicolumn{2}{c}{\textbf{Object HalBench}} & \textbf{MathVista}    \\
  &   CHAIR$_S$$\downarrow$ & CHAIR$_I$$\downarrow$ & Acc$\uparrow$    \\

\midrule
PixelReasoner-7B & 22.0 & 7.8 & 62.6 \\

\midrule

\multicolumn{4}{l}{\textbf{\textit{Intervention Location:}}} \\

All Thinking & 16.1 & 5.7 & 61.4  \\
Normal-only & 21.8 & 7.7 & 62.7 \\
Divergent-only (Ours) & \textbf{15.4} & \textbf{5.3} & \textbf{64.0} \\

\midrule
\multicolumn{4}{l}{\textbf{\textit{Intervention Technique:}}} \\

PAI~\cite{liu2024paying} & 19.6 & 7.0 & 63.2 \\
MemVR~\cite{memvr} & 19.1 & 6.8 & 63.4 \\
\textit{Visual Entropy} (Ours) & \textbf{15.4} & \textbf{5.3} & \textbf{64.0} \\

\bottomrule
\end{tabular}
\end{table}

\subsection{Different Model Settings}
\label{Different Model Settings}
In Table~\ref{tab:different_model_settings}, we explore several alternative settings for our proposed method. The related analyses are conducted on Object HalBench~\cite{chair} and MathVista~\cite{mathvista} with PixelReasoner-7B~\cite{pixelreasoner} as the backbone.

\noindent
\textbf{Intervention Mode.} 
Based on findings from Sec.~\ref{sec:finding2}, we only modulate the reasoning process when model is performing divergent thinking (denoted by high \textit{visual entropy}). To validate this approach, we compare it against two alternative strategies. As shown in the Table~\ref{tab:different_model_settings}, “All Thinking” denotes adjusting the entire reasoning process of MCoT models, and “Normal-only” means intervening only during the normal thinking steps. The results demonstrate that both alternatives yield inferior performance compared to our current setting.

\noindent
\textbf{Intervention Technique.}
In the bottom part of Table~\ref{tab:different_model_settings}, we further evaluate the proposed approach by comparing it with several alternative strategies. In particular, during divergent thinking steps, we replace our intervention strategy with PAI~\cite{liu2024paying} and MemVR~\cite{memvr}. The results show that our approach outperforms these alternatives, validating the effectiveness of our method.

\begin{table}[t]
\centering
\setlength{\tabcolsep}{4pt}
\caption{Compatibility with other mitigation strategies.}
\label{tab:combine_use}
\begin{tabular}{lccc}
\toprule
\multicolumn{1}{l}{\multirow{2}{*}{\textbf{Methods}}}  &   \multicolumn{2}{c}{\textbf{Object HalBench}} & \textbf{MathVista}    \\
  &   CHAIR$_S$$\downarrow$ & CHAIR$_I$$\downarrow$ & Acc$\uparrow$    \\

\midrule
PixelReasoner-7B & 22.0 & 7.8 & 62.6 \\

\midrule
+ DoLa~\cite{chuang2023dola} & 20.6 & 7.1 & 62.2  \\
+ DoLa + Ours & \textbf{17.9} & \textbf{6.0} & \textbf{63.4} \\

\midrule
+ VCD~\cite{vcd} & 19.8 & 7.2 & 61.9 \\
+ VCD + Ours & \textbf{16.1} & \textbf{5.3} & \textbf{63.5} \\

\bottomrule
\end{tabular}
\end{table}

\begin{table}[t]
\centering
\setlength{\tabcolsep}{4pt}
\caption{Compatibility with different decoding strategies.}
\label{tab:different_decoding}
\begin{tabular}{lccc}
\toprule
\multicolumn{1}{l}{\multirow{2}{*}{\textbf{Methods}}}   &   \multicolumn{2}{c}{\textbf{Object HalBench}} & \textbf{MathVista}    \\
  &   CHAIR$_S$$\downarrow$ & CHAIR$_I$$\downarrow$ & Acc$\uparrow$    \\

\midrule

\multicolumn{1}{l}{\textbf{\textit{Greedy Decoding:}}} \\
Original & 22.0 & 7.8 & 63.1 \\
+ Ours & \textbf{15.4} & \textbf{5.3} & \textbf{63.6} \\

\midrule
\multicolumn{1}{l}{\textbf{\textit{Nucleus Sampling:}}} \\
Original & 25.8 & 9.4 &  62.6 \\
+ Ours & \textbf{21.5} & \textbf{7.2} & \textbf{64.0} \\

\midrule
\multicolumn{1}{l}{\textbf{\textit{Beam Search:}}} \\
Original & 20.8 & 7.4 & 62.8 \\
+ Ours & \textbf{15.1} & \textbf{4.9} & \textbf{64.1} \\

\bottomrule
\end{tabular}
\end{table}

\subsection{Compatibility Analyses}

\noindent
\textbf{Compatibility with Hallucination Mitigation Methods.}
\label{para:compatible_with_hallu}
We further investigate whether our method is compatible with traditional hallucination mitigation techniques, \textit{e.g.,} DoLa~\cite{chuang2023dola} and VCD~\cite{vcd}.
As shown in Table~\ref{tab:combine_use}, it can be found that while DoLa and VCD effectively reduce 
hallucinations of the PixelReasoner-7B~\cite{pixelreasoner}, they 
are also harmful for the model's math reasoning capability. For instance, VCD degrades performance on MathVista by 0.7 points.
In contrast, applying our method with these hallucination mitigation techniques not only reduces hallucinations but also boosts the model's reasoning capacity. 
These results show that our method is highly compatible with different hallucination mitigation methods.

\noindent
\textbf{Compatibility with Different Decoding Methods.}
To assess the robustness of our strategy to different decoding manners, we conduct experiments on PixelReasoner-7B~\cite{pixelreasoner} using greedy decoding~\cite{germann2003greedy}, nucleus sampling~\cite{holtzman2019curious}, and beam search~\cite{steinbiss1994improvements}. As shown in Table~\ref{tab:different_decoding}, we observe that our method yields consistent improvements across different decoding strategies. For instance, under beam search, our method reduces CHAIR$_S$ from 20.8 to 15.1 and CHAIR$_I$ from 7.4 to 4.9. More importantly, our method also enhances the model's reasoning capacity across all tested decoding methods.

\begin{figure}[t]
  \centering
  \includegraphics[width=1\linewidth]{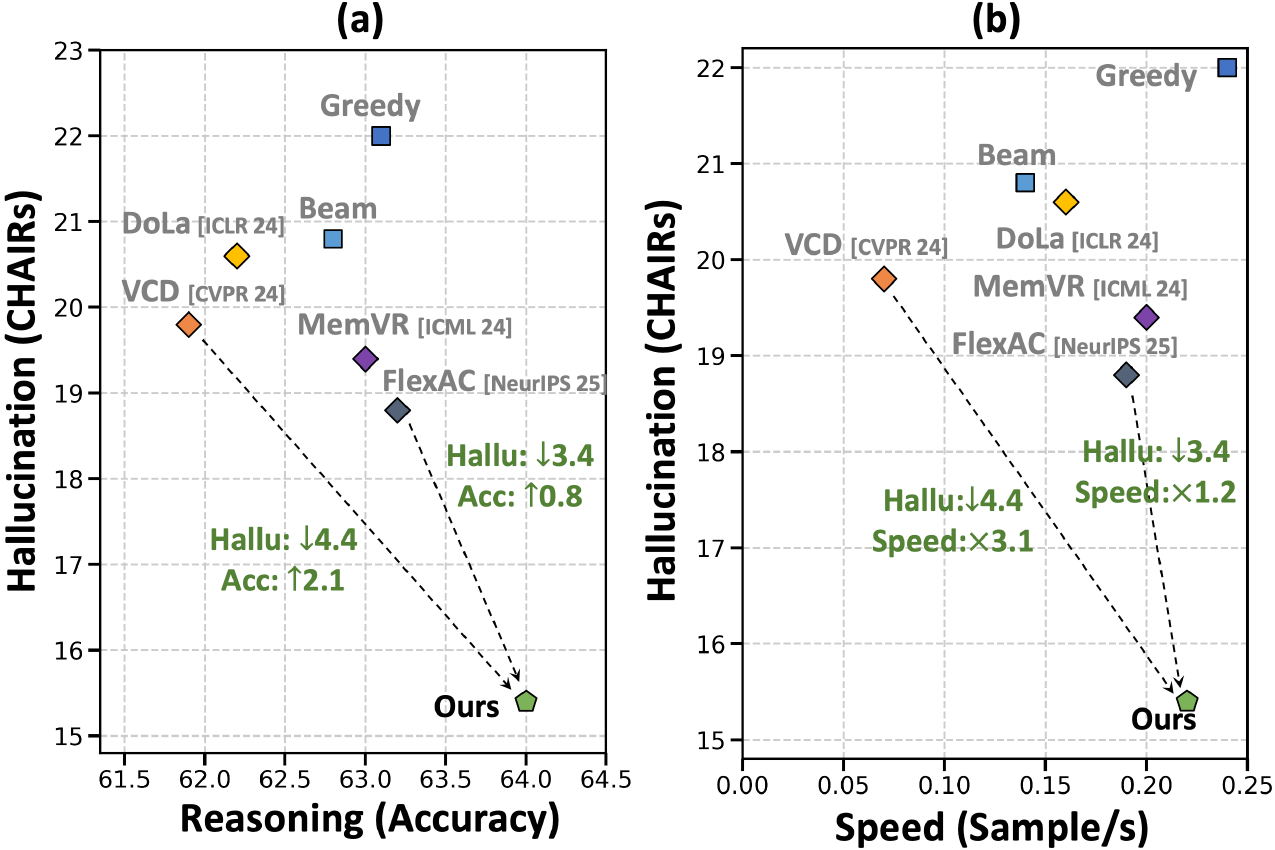}
  \caption{(a) Performance balance between reasoning and hallucinations. (b) Trade-off between effectiveness and efficiency.}
  \label{fig:comparison}
\end{figure}

\begin{figure}[t]
  \centering
  \includegraphics[width=1\linewidth]{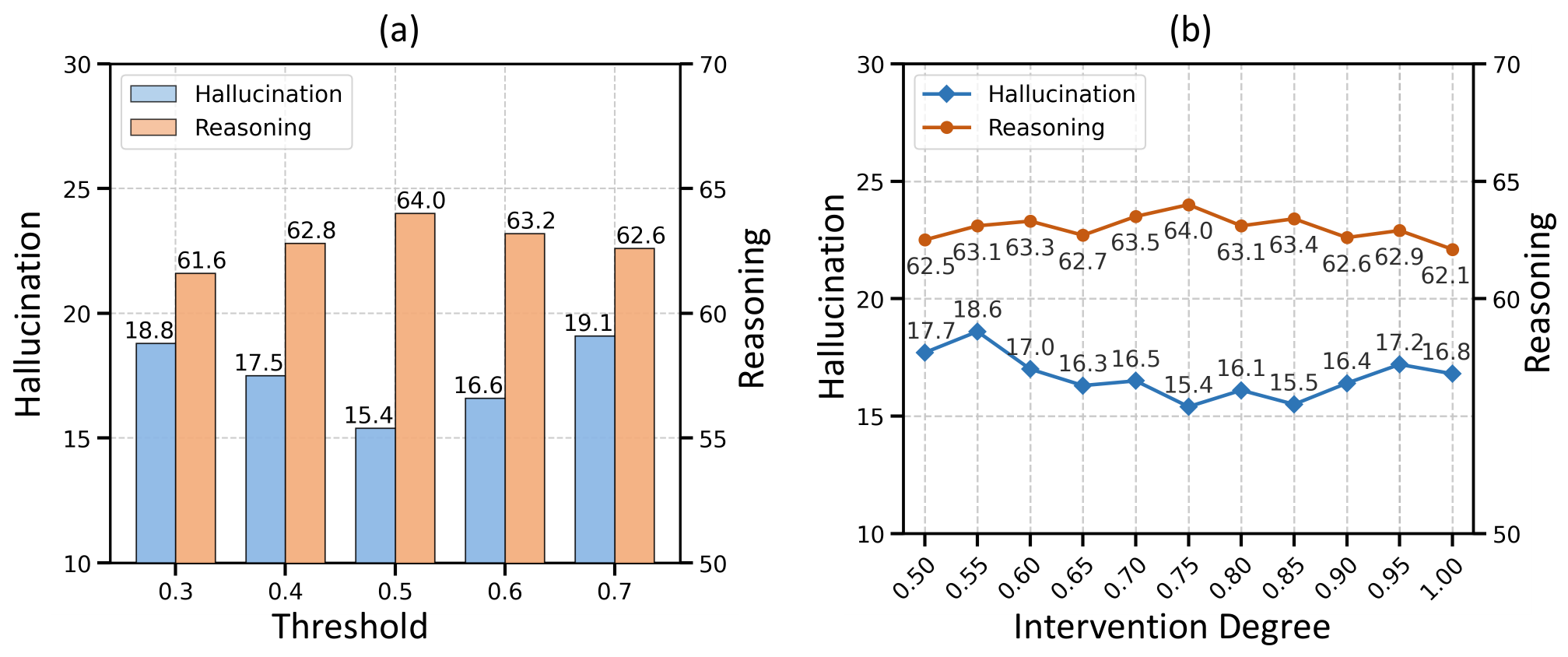}
  \caption{(a) Effects of different threshold $\gamma$. (b) Effects of different intervention degree $\alpha$.}
  \label{fig:hyper}
\end{figure}

\subsection{More Results and Analyses}
\label{sec:5.5}

\textbf{Balance between Reasoning and Hallucinations.}
In Fig.~\ref{fig:comparison}~(a), we use MathVista~\cite{mathvista} and Object HalBench~\cite{chair} to evaluate the balance between reasoning and hallucination performance. From this figure, it can be observed that our method achieves the optimal trade-off.
For example, compared to VCD~\cite{vcd}, our approach not only improves accuracy by 2.1\% but also reduces hallucinations by 4.4 points. The results show that our method effectively mitigates hallucinations without harming their reasoning performance.

\noindent
\textbf{Trade-off between Effectiveness and Efficiency.}
In Fig.~\ref{fig:comparison}~(b), we compare the inference speed of our strategy with several hallucination mitigation methods. The results indicate that our approach incurs only a marginal latency overhead compared to greedy decoding. Moreover, compared to FlexAC~\cite{FlexAC}, the proposed method not only achieves lower hallucination ratios but is also 1.2$\times$ faster.

\noindent
\textbf{Hyperparameter Settings.}
In Fig.~\ref{fig:hyper}~(a), we evaluate the impact of different thresholds $\gamma$. We observe that our \textit{visual entropy} achieves the optimal performance when this value is 0.5. For intervention degree $\alpha$, as shown in Fig.~\ref{fig:hyper}~(b), the optimal performance can be obtained with $\alpha=0.75$.
Notably, although we conduct the above experiments on PixelReasoner-7B, further investigation using different backbones shows that $\gamma = 0.5$ and $\alpha = 0.75$ still yield satisfactory performance for other models, including GRIT-3B and R1-Onevision-7B. 
These results suggest that hyperparameter choices are model-agnostic, avoiding labor-intensive hyperparameter tuning when adapting our method to different models. 

\begin{figure}[t]
  \centering
  \includegraphics[width=1\linewidth]{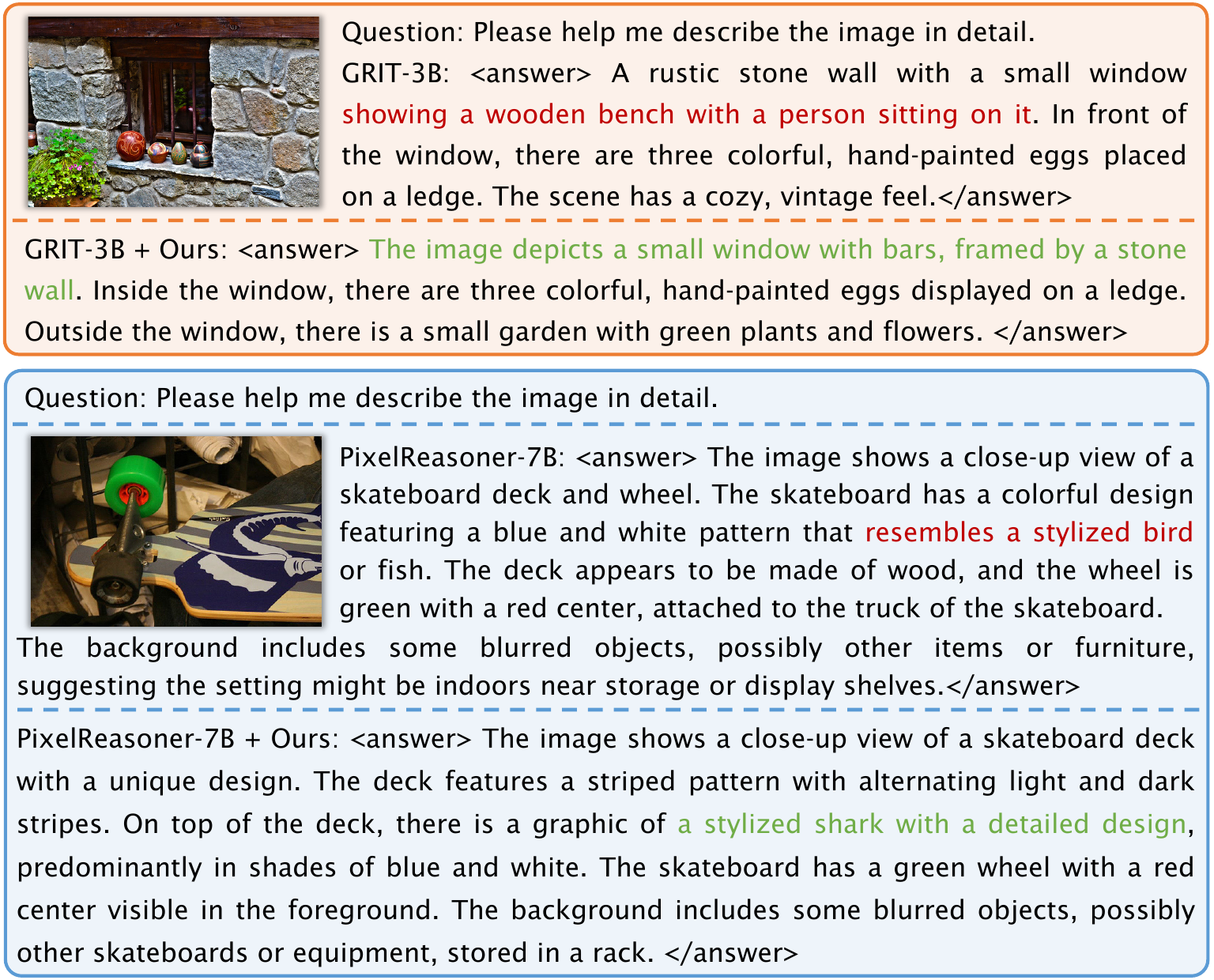}
  \caption{Qualitative analyses. 
  The original MCoT models produce hallucinated contents. Our method effectively mitigates hallucinations and performs faithful reasoning and answering.}
  \label{fig:qual}
\end{figure}

\subsection{Qualitative Evaluation}

In Fig.~\ref{fig:qual}, we provide qualitative evaluations of our method on two prevalent MCoT models.
The results show that the original MCoT models suffer from serious hallucination problems.
In contrast, our method effectively mitigates hallucinations, producing more faithful answers while preserving the original reasoning capacity of MCoT models.

%% file: sec/6_conclusion.tex
\section{Conclusion}
\label{sec:conclusion}

In this paper, we first explore the hallucination occurrence patterns in Multimodal Chain-of-Thought (MCoT) models. Through comprehensive experiments, we find that a key factor for
hallucination emergence is \textit{divergent thinking}. Further investigations demonstrate that the introduced \textit{visual entropy} can be utilized to effectively localize divergent thinking steps and mitigate hallucinations. Based on these insights, we propose a simple yet effective strategy that can adaptively adjust the decoding process using entropy values. 
We expect that our work can provide a general and interpretable approach to alleviate hallucination challenges for the MCoT paradigm.  

\clearpage